\documentclass[journal]{IEEEtai}

\usepackage[colorlinks,urlcolor=blue,linkcolor=blue,citecolor=blue]{hyperref}

\usepackage{color,array}

\usepackage{graphicx}

\usepackage{bbm}
\usepackage{amsmath}
\usepackage{multirow}
\usepackage{amssymb}
\usepackage{fontawesome}

\begin{document}

\title{Rethinking Multiple Instance Learning: Developing an Instance-Level Classifier via Weakly-Supervised Self-Training}

\author{Yingfan~Ma\textsuperscript{*},
Xiaoyuan~Luo\textsuperscript{*},
Mingzhi~Yuan\textsuperscript{*},
Xinrong~Chen\textsuperscript{\faEnvelopeO},
Manning~Wang\textsuperscript{\faEnvelopeO}
\thanks{Yingfan Ma, Xiaoyuan Luo, Mingzhi Yuan, Xinrong Chen and Manning Wang are with the Digital Medical Research Center, School of Basic Medical Science, Fudan University, Shanghai 200032, China, and also with the Shanghai Key Laboratory of Medical Image Computing and Computer Assisted Intervention, Shanghai, China, 200032.}
\thanks{* These authors have contributed equally to this work.}
\thanks{{\faEnvelopeO} Corresponding author.}
\thanks{E-mail: 22211010089@m.fudan.edu.cn, \{xyluo19, mzyuan20, chenxinrong, mnwang\}@fudan.edu.cn}}


\maketitle

\begin{abstract}
    Multiple instance learning (MIL) problem is currently solved from either bag-classification or instance-classification perspective, both of which ignore important information contained in some instances and result in limited performance. 
  For example, existing methods often face difficulty in learning hard positive instances. 
  In this paper, we formulate MIL as a semi-supervised instance classification problem, so that all the labeled and unlabeled instances can be fully utilized to train a better classifier. 
  The difficulty in this formulation is that all the labeled instances are negative in MIL, and traditional self-training techniques used in semi-supervised learning tend to degenerate in generating pseudo labels for the unlabeled instances in this scenario. 
  To resolve this problem, we propose a weakly-supervised self-training method, in which we utilize the positive bag labels to construct a global constraint and a local constraint on the pseudo labels to prevent them from degenerating and force the classifier to learn hard positive instances. 
It is worth noting that easy positive instances are instances are far from the decision boundary in the classification process, while hard positive instances are those close to the decision boundary.
  Through iterative optimization, the pseudo labels can gradually approach the true labels. Extensive experiments on two MNIST synthetic datasets, five traditional MIL benchmark datasets and two histopathology whole slide image datasets show that our method achieved new SOTA performance on all of them. 
  The code will be publicly available.  
\end{abstract}
\begin{IEEEImpStatement}
MIL is a primary research field within weakly supervised learning and has numerous applications, such as histopathology analysis, video anomaly detection, and drug design. However, current mainstream methods in MIL predominantly rely on bag labels to train the network, consequently overlooking the potential of instance-level supervision information. In our paper, we highlight the information loss inherent in traditional bag-based methods and introduce a novel weakly-supervised self-training framework that fully harnesses the available supervision information. Comprehensive experiments demonstrate that our framework achieves significant improvements across multiple datasets. Furthermore, our approach holds promise for applications in other MIL domains.
\end{IEEEImpStatement}


\section{Introduction}
\IEEEPARstart{M}{ultiple} instance learning (MIL) is a weakly supervised learning problem encountered in many fields \cite{Carbonneau_2018, Cheplygina_2015}, such as drug design \cite{ref29}, pathological image classification \cite{ref1,ref2,ref7,ref14, TAI} and video anomaly detection \cite{ref30,ref31}. 
In the typical MIL scenario, multiple instances form a bag whose label is known for training, whereas the label of instances is unknown.
Traditional MIL methods mainly focus on bag-level classification, but instance-level classification is also important. It can help improve the interpretability of the classification model and discover knowledge. For example, it is crucial for localizing positive regions in pathological images.

\begin{figure}[ht]
\begin{center}
\centerline{\includegraphics[width=1.0\columnwidth]{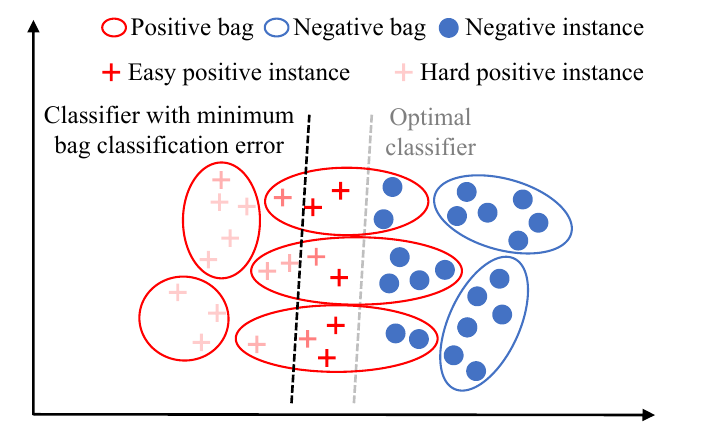}}
\caption{
    Illustration of the problem of bag-classification approach in MIL. 
    In the 2D instance feature space, the easy positive instances are far away from the negative instances, whereas the hard positive instances are near the negative instances. 
    For bag-classification methods, the minimum bag classification error can be achieved when the classifier learns to distinguish a small number of easy positive instances, and all the instance information is not fully utilized. The goal of further optimization is learning hard positive instances.
}
\label{fig1}
\end{center}
\vspace{-1em}
\end{figure}

Since only bag labels are available for training, most existing mainstream approaches address the problem from the perspective of bag classification. 
This bag-classification approach aggregates the features of all instances in a bag to obtain bag-level features, and trains a bag-level classifier to classify them. 
In recent years, the research focus in this direction has been to develop more effective aggregation functions by using attention \cite{ref1,ref3,ref4}, graph convolution network (GCN) \cite{ref5} and Transformer \cite{ref6}. 
However, bag classification approach compromises the instance-level classification ability because the loss function is mainly built on the bag classification results, and some important instance-level information is ignored. 
As shown in Figure \ref{fig1}, in the instance feature space, the bag-level classifier can correctly classify a bag even when it can only distinguish one easy positive instance in it, lacking the motivation for further optimization. 
This creates two drawbacks: (1) Insufficient instance classification capability. The model may not be able to distinguish the hard positive instances and thus has insufficient instance-level classification ability. 
 (2) Poor do not have labels for potentially positive instancesgeneralization. If all positive training bags have some easy positive instances, the trained model may not generalize well on bags with only hard positive instances. 
Furthermore, we theoretically analyze the bag-classification approach from the perspective of information theory, and show that, regardless of the aggregation methods, this approach will lose a large amount of information inside negative bags.

Besides the bag-classification approach, there are some studies that solve the MIL problem by building an instance-level classifier. 
This instance-classification approach first directly trains an instance-level classifier, and then aggregates the instance classification scores in each bag to obtain the bag classification result. 
However, we only know the labels of some negative instances, which are from the negative bags, and do not have labels for potentially positive instances. 
Therefore, the existing instance-classification methods either treat all instances in the positive bag as positive \cite{ref13}, or select a small number of key instances from positive bags as positive through top-k or other methods \cite{ref2,ref10}. 
Obviously, the former approach will introduce many noisy labels, and the latter will lose most of the instance information in the positive bags.

In this paper, we formulate traditional binary classification MIL as a semi-supervised instance classification problem to utilize all instance-level information more efficiently.  Motivated by semi-supervised learning, which leverages a small amount of labeled data along with unlabeled data to enhance model performance,
we formulate MIL as an instance classification problem, the labels of some instances (instances in all negative bags) are known, while the labels of the other instances (instances in positive bags) are unknown, which is similar to a typical semi-supervised learning setting. 
The only difference is that in MIL all the known labels are negative, which makes it difficult to use general self-training techniques in semi-supervised learning to learn from the unlabeled data. 
Self-training usually starts from training a model with available labeled data. Then, the trained model is used to generate pseudo labels for the unlabeled data, which is called self-labeling, so that the unlabeled data together with their pseudo labels can be used in model optimization in an iterative way. 
However, the known instance labels in the MIL problem are all negative. It is impossible to train an initial classifier with all-negative instances, and iteratively optimizing the classifier and the pseudo labels only with the guidance of the true negative labels will make both the model and the pseudo labels fall into degeneration.

To resolve this problem, we propose a weakly-supervised self-training method. 
Specifically, we apply a global constraint and a local constraint derived from the weak supervision information of positive bag labels in the self-labelling process, so that we can prevent the pseudo labels and the trained network from degenerating. 
Furthermore, the global constraint also forces the network to learn hard positive instances. 
Through iteratively alternating between self-labelling and model updating under these two constraints, the pseudo labels can gradually approach the true labels and the model also gradually learns to classify the instances correctly. 
Formulating MIL as a special semi-supervised learning problem and solving it using the proposed weakly-supervised self-training method enables efficient utilization of instance-level information and outputs a successful instance-level and bag-level classifier.

Although MIL problem is formulated as a SSL problem in some traditional machine learning studies \cite{ref50,ref51,ref52},  we apply a similar framework in a more efficient way by integrating advanced techniques, especially with the great power of newly proposed self-training frameworks.

We conducted extensive experiments on two MNIST synthetic datasets, five traditional MIL benchmark datasets and two real pathological image datasets, CAMELYON 16 and TCGA, and the results show that our method outperforms existing methods by a large margin.

Overall, the main contributions of this paper are summarized as follows:
\begin{itemize}
\item We theoretically analyze the problems of solving MIL with traditional bag-classification approach, including the hard positive instances recognition problem in instance-level classification and the generalization problem in bag-level classification. 
\item We reformulate MIL as a special semi-supervised instance classification problem, and propose a weakly-supervised self-training method to solve the problem. Our method trains an instance classifier in an end-to-end manner and achieves better information utilization. 
\item We conduct extensive experiments. First, we build two synthetic datasets to address the drawbacks of the traditional methods and to verify the effectiveness of our method. Then, we further perform experiments on five traditional MIL benchmark datasets and two real pathological image datasets to demonstrate the superiority of our method.
\end{itemize}

\section{Related work}
\subsection{Multiple instance learning}
Multiple instance learning (MIL) is a typical weakly supervised learning scheme where a label is associated with a group of instances, which is usually called as a bag, and instance-level labels are not provided in the training data. 
The primary objective of MIL is bag-level classification, while instance-level classification is also very important in many applications like whole slide images processing and video anomaly detection. 
Existing methods solve the MIL problem from one of the two perspectives: bag-classification approach and instance-classification approach. 
The former one aggregates the instances maintaining bag-level information and then build a bag-level classifier, and the latter one build instance-level classifier and aggregate the instance prediction scores to obtain bag-level prediction \cite{CHEPLYGINA2019280}. 

For bag-classification approaches, how to aggregate instance features within a bag to generate the bag features plays a central role. Traditional aggregation methods typically explore handcrafted permutation-invariant aggregation functions such as max pooling, mean pooling, quantile aggregation and dynamic aggregation \cite{ref13,ref8,ref9,ref26}, but they are inflexible and not robust, easily leading to overfitting \cite{ref3}. 
Recently, aggregation operators parameterized with neural networks achieved better results. Ilse et al. introduced an attention network in aggregation \cite{ref4} and Li et al. proposed a nonlocal aggregation method based on the attention mechanism \cite{ref1}. 
Some researchers further employed RNN \cite{ref7}, GNN \cite{ref5,ref27} and Transformer \cite{ref6} as feature aggregator and achieved better results than vanilla attention network. 
However, the aggregation operation may lose some information in the instances and thus compromise the bag and instance classification ability. 

For instance-classification approaches, most current methods select some key instances from positive bags and treat them as positive to train the instance classifier, because the positive instances are unknown. Chikontwe et al. trained a network to first predict the positive probability of each patch in positive bags and then selected the top-k instances as positive instances \cite{ref2}. 
Chen et al. trained a focal-aware module separately to select key instances in a multiscale manner \cite{ref10}. 
Unfortunately, it is difficult to choose all potentially positive instances from positive bags because the number of positive instances in positive bags is variable and may be small. 
In addition, most strategies only use selected key instances to train the classifier and cannot benefit from other information in positive bags.
More recently, WENO \cite{ref49} avoid these problems through knowledge distillation but still relies on the bag classifier. 

\subsection{Semi-supervised learning and self-training}
Semi-supervised learning is a branch of machine learning that deal with the problem of how to train a model with a small amount of labeled data and a large amount of unlabeled data. 
There exist many mature approaches in the semi-supervised field \cite{ref46}, among which self-training-based methods have recently received extensive research attention and achieved SOTA performance on multiple semi-supervised tasks.

SSL methods based on self-training are mainly divided into two types. 
The first type explicitly generates pseudo labels for unlabeled data, and then uses pseudo labels and real labels to jointly train the network \cite{ref35,ref36}. 
MixMatch \cite{ref40}, ReMixMatch \cite{ref41} and FixMatch \cite{ref42} belong to this paradigm, and they use an online method to generate pseudo labels. 
SLA \cite{ref43} models the pseudo label generation process with optimal transfer, and PAWS \cite{ref44} combines contrastive self-supervised learning with self-training. 
We also utilized optimal transfer in generating pseudo labels, but our proposed method is different from SLA in that we utilize weak-constraints in MIL to make them more informative. 
The second type of is also known as knowledge distillation, with alternative approaches Mean Teacher \cite{ref37}, Noise Student \cite{ref38}, MPL \cite{ref39}. More details about knowledge distillation can be found in \cite{ref47}. 

\section{Method}
\subsection{Problem formulation}
We address traditional binary multiple instance learning (MIL) problem, where a bag ${X}_{i}$ is composed of a set of $K$ instances ${X}_{i}=\left\{x_{i, 1}, x_{i, 2}, \ldots, x_{i, K}\right\}$, and the bag label ${Y}_{i}$ is available but the instance labels $\left\{y_{i, 1}, y_{i, 2}, \ldots, y_{i, K}\right\}$ are unknown.
At the same time, the following relationship exists between the bag label and the instance labels:
\begin{equation}
    \label{eq1}
    {Y}_{i}= \begin{cases}0, & \text { iff } \sum_{k} y_{i, k}=0 \\ 1, & \text { else }\end{cases}
\end{equation}

The basic task in MIL is bag-level classification, and a more challenging task is instance-level classification, which is also an important objective in many applications. 

\subsection{Bag-classification approach}\label{sec2.2}
We first briefly introduce the current mainstream bag-classification approach in solving MIL problem so that we can clearly analyze their limitations in the next subsection. 
This type of methods first encodes each instance into a feature vector through an encoder $f$, and then all instances in a bagaccording to their dissimilarity to are aggregated through an aggregation function $g$ to obtain the bag features, which are processed by a classifier $\varphi$ to predict the bag label:
\begin{equation}
  \widehat{Y}_{i}=(\varphi \circ g)\left(f\left(x_{i, 1}\right), \ldots, f\left(x_{i, K}\right)\right)
\end{equation}

The whole model is trained with the cross entropy between the prediction of all bags $\widehat{Y}_{i}$ and the known bag labels $Y$:
\begin{equation}
  {\min } CE(Y, \widehat{Y})
\end{equation}

The key component in bag-classification approach is the aggregation function $g$, and most recent methods build $g$ on the attention mechanism \cite{ref1,ref3,ref4}:
\begin{equation}
  g\left(f\left(x_{i, 1}\right), \ldots, f\left(x_{i, K}\right)\right)=\sum_{k=1}^{K} \alpha_{i, k} f\left(x_{i, k}\right)
\end{equation}
\begin{equation}
  \alpha_{i, k}=\frac{\exp \left\{w^{T} \tanh \left(V f\left(x_{i, k}\right)^{T}\right)\right\}}{\sum_{j=1}^{K} \exp \left\{w^{T} \tanh \left(V f\left(x_{i, j}\right)^{T}\right)\right\}}
\end{equation}
where $\alpha_{i, k}$ is the attention score predicted by the attention network which is parameterized by $w$ and $V$. $ $w$ \in \mathbb{R}^{L \times 1}$ and $  $V$ \in \mathbb{R}^{L \times M}$ are learnable prameters.

For instance-level classification, the attention scores can be used to do instance-level prediction after normalization:
\begin{equation}
  \hat{y}_{i, k}=\operatorname{norm}\left(\alpha_{i, k}\right)
\end{equation}

\subsection{Insufficient instance classification and hard positive instance learning problem}\label{sec2.3} 
For easier understanding the defects of the bag-classification approach, we consider a simple example shown in Figure \ref{fig1}, where all instances have been encoded into a 2D feature vector and they are linearly separable in the feature space. 
The positive instances are further divided into easy positive instances and hard positive instances according to their distance to the negative instances. Since the bag-classification approach builds the loss function mainly on the bag classification results, it is possible that the network stop optimizing after it only learns to distinguish a small number of positive instances in each positive bag. 
This phenomenon will lead to two problems: (1) The instance-level classification ability of the network is insufficient, because it only needs to learn to distinguish a part of the positive instances to identify the positive bags and may lose the motivation to distinguish the other positive instances; (2) Poor generalization ability of bag-level classification because of insufficient hard positive instance learning. 
When there are both easy positive and hard positive instances in positive bags of the training data, the network will tend to learn to distinguish the easy positive instances to predict the bag labels, so it will be unable to distinguish the bags composed of only hard positive instances.

We further analyze the MIL problem from the perspective of information theory and derive the following Proposition 1. 
The proof is in the Appendix \ref{proof}. 
Intuitively, if grouping instances into bags is regarded as an operation on the instances, the proposition states that the instance labels have the maximum information entropy, and the instance aggregation operation reduces the information in the training data.
We calculate and visualize the entropy difference between the instance labels and the bag label in the Appendix \ref{figure}. 
We conclude that the information loss increases when the bag-length $K$ increases. 
As a bag usually contains thousands of instances in scenarios like pathological image analysis, the bag classification approach will lose a lot of information and is not the best choice for instance-level classification. Therefore, instance-level methods can preserve more information when handling large-scale instance data and have the potential for more precise classification.

\textbf{Proposition 1:} Given $K>1$ independent and identically distributed instances, the probability of each instance $x_i$ being positive ranges from 0 to 1, and the information entropy of these instances is $H_I$. 
After the instance aggregation operation is performed on the $K$ instances, and the information entropy of the bags is $H_B$, then $H_{B}<H_{I}$.

\begin{figure}[!htbp]
  \begin{center}
  \centerline{\includegraphics[width=1.0\columnwidth]{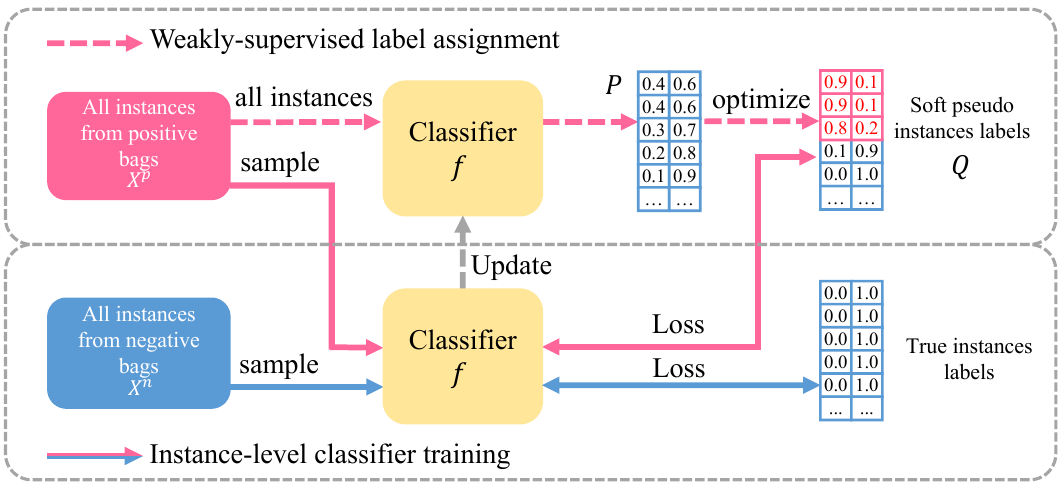}}
  \caption{
      The framework of the proposed MIL-SSL method, in which we formulate multiple instance learning (MIL) as a semi-supervised learning (SSL) instance classification problem. 
      Our method iteratively alternates between two steps: (1) weakly-supervised pseudo label assignment, in which we use the current instance classifier to predict all unlabeled instances from positive bags to obtain $P$ and then calculate the pseudo labels $Q$ by solving an optimization problem under global and local constraints; 
      (2) instance classifier training, in which we train the instance classifier with both the instances from negative bags and their true labels and the instances from positive bags and their pseudo labels.}
  \label{fig3}
  \end{center}
\end{figure}

\begin{figure*}[!htbp]
    \begin{center}
    \centerline{\includegraphics[width=2.0\columnwidth]{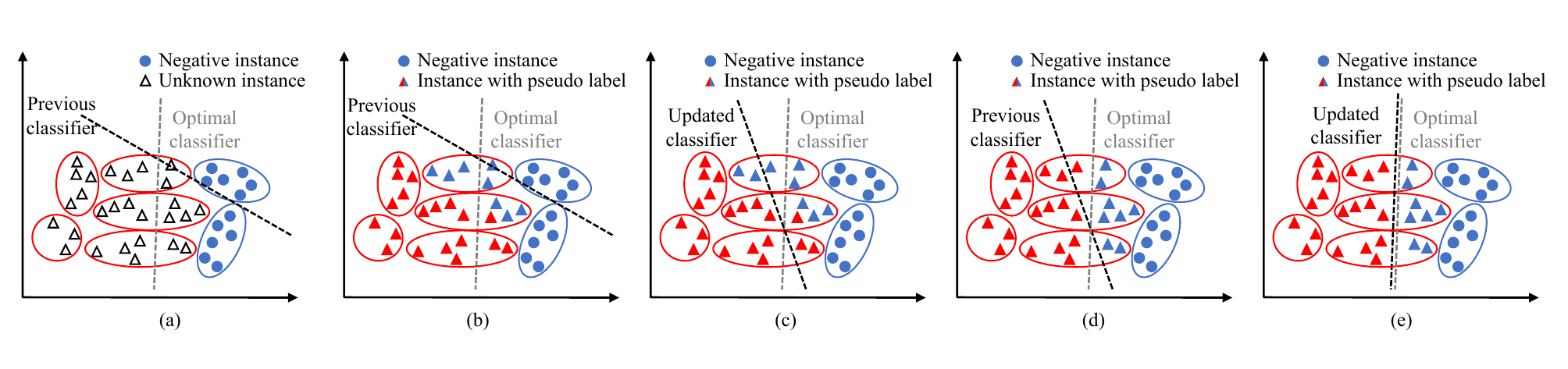}}
    \caption{
      Illustration of the iterative updating of the pseudo labels and the instance-level classifier in the proposed method. 
      (a) Initially, the instance labels in all negative bags are known to be negative but the instance labels in positive bags are unknown, and the instance-level classifier is randomly initialized; (b) pseudo labels for all instances in positive bags are generated based on the prediction results of instance-level classifier, and the pseudo labels are constrained to obey a certain distribution; 
      (c) the pseudo labels and true negative labels in the negative bag are used to further train the instance-level classifier, and (d) then new pseudo labels are generated with the new classifier; 
      (e) the pseudo labels gradually approach real labels and the instance-level classifier gradually approaches the optimal classifier. 
    }
    \label{fig4}
    \end{center}
  \end{figure*}

\subsection{Our method}
As analyzed in section \ref{sec2.2} and \ref{sec2.3}, more information can be retained if we directly train a model using all the instances in a MIL problem. 
Therefore, we formulate traditional binary MIL as an instance-level semi-supervised learning (SSL) problem so that all the instances in both negative bags and positive bags can be fully utilized to train an instance-level classifier. 

In semi-supervised learning, self-training is often used to learn from unlabeled data. In a typical self-training process, a classifier is initially trained with the labeled data. 
Then self-training iteratively alternates between the following two steps: use the current classifier to generate labels for the unlabeled data, which are called pseudo labels, and update the classifier using both the originally labeled data and the data with pseudo labels. 
During this process, both the pseudo labels and the classifier are gradually optimized. 
However, since all the known labels in MIL are negative, directly adopting the above self-training strategy will make the classifier fall into degeneration and classify all instances into negative. 
In this paper, we propose weakly-supervised self-training to prevent the classifier from falling into degeneration. 
As shown in Figure \ref{fig3}, our proposed self-training method also consists of two alternative steps: 
(1) Weakly-supervised label assignment. 
In addition to use current classifier to predict the labels $P$ for all the unlabeled instances, which are from the positive bags, we solve a constrained optimization problem to assign the pseudo labels $Q$ on the base of $P$. 
The constraints of the optimization problem are derived from the weak supervision of the positive bags and the constrained optimization problem is solved with optimal transport algorithm; 
(2) Instance-level classifier training: the instance classifier is trained using both the instances with pseudo labels and the instances with true negative labels from the negative bags. 
The two steps will be introduced in detail in the following two subsections.

\subsubsection{Weakly-supervised label assignment}
In this step, new pseudo labels are generated for all unlabeled instances, which are the instances from the positive bags. 
We first use the current instance-level classifier $f$ to predict the probability that these instances are positive:
\begin{equation}
    \label{eq2}
    p_{i, j}^{p}=f\left(x_{i, j}^{p}\right)
\end{equation}
where the subscripts $i$ and $j$ are the bag and instance index, respectively, and the superscript $p$ indicates that the instance belongs to a positive bag. 
Similarly, we use superscript $n$ to indicate that an instance comes from a negative bag. 
We arrange the prediction results of all the unlabeled instances as $P \in \mathbb{R}^{N \times 2}$. 
Each column of $P$ corresponds to the prediction of an instance, and the first-column and the second-column elements are the probability of the instance being positive and negative, respectively, and they sum up to one. 
In the MIL scenario, since the known labels are all negative, if the predicted results are directly taken as pseudo labels, the network will tend to fall into a degenerate solution:
\begin{equation}
    \label{eq3}
    p_{i, j}^{p}=0, y_{i, j}^{n}=0, f\left(x_{i, j}\right)=0
\end{equation}
where $y_{i, j}^{n}$ is the true label of the $j$th instance from the $i$th negative bag. 
To prevent degeneration, we then construct pseudo label matrix $Q \in \mathbb{R}^{N \times 2}$ from $P$ by solving a constrained optimization problem. 
Concretely, we minimize the distance between $Q$ and $P$ under a global constraint that $Q$ must obey a specific marginal distribution:
\begin{equation}
    \label{eq4}
    \min _{Q \in U(r, c)}\langle Q,-\log P\rangle
\end{equation}
\begin{equation}
    \label{eq5}
    r=\left[\begin{array}{c}
    \mu \cdot N \\
    (1-\mu) \cdot N
    \end{array}\right], \quad c=\mathbbm{1}
\end{equation}
where $\langle\cdot\rangle$ represents the Frobenius dot-product, $r$ and $c$ are the marginal projections of the matrix $Q$ onto its rows and columns, $U(r, c)$ represents the transport polytope with marginal distribution $r$ and $c$, $\mu$ represents the percentage of positive instances out of all unlabeled instances and $\mathbbm{1}$ is the vector of all ones of appropriate dimensions. 
This constrained optimization problem is an optimal transport problem and the fast Sinkhorn-Knopp algorithm \cite{ref12} is applied to solve it. 
First, a regularization term is introduced into the objective:
\begin{equation}
    \label{eq6}
    \min _{Q \in U(r, c)}\langle Q,-\log P\rangle+\frac{1}{\lambda} \mathrm{KL}\left(Q \| r c^{T}\right)
\end{equation}
where $\mathrm{KL}$ represents the Kullback-Leibler divergence. 
When $\lambda$ is very large, optimizing Equation \ref{eq6} is of course equivalent to optimizing \ref{eq4}. 
The introducing of the regularization term leads to a solution:
\begin{equation}
    \label{eq7}
    Q=\operatorname{diag}(\alpha) P^{\lambda} \operatorname{diag}(\beta)
\end{equation}
where $\alpha$ and $\beta$ are two scaling coefficients, and can be obtained via matrix scaling iteration:
\begin{equation}
    \label{eq8}
    \forall y: \alpha_{y} \leftarrow\left[P^{\lambda} \beta\right]_{y}^{-1}, \forall i: \beta_{i} \leftarrow\left[\alpha^{T} P^{\lambda}\right]_{i}^{-1}
\end{equation}

The reason for constraining the probability of positive instances in all positive bags instead of in each bag separately is that the probability of positive instances in each bag may vary greatly in real MIL problemsin real MIL problems. However, the percentage of all positive instances out of all positive bags is relatively stable and can be easily estimated from training data or can be tuned as a hyperparameter. 
The global constraint not only prevents the pseudo labels from degeneration, but also forces the classifier to learn both easy and hard positive instances. 
However, the global constraint does not fully utilize the weak supervision information that there must exist at least one positive instance in each positive bag. 
Therefore, we further design a local constraint for the pseudo labels. Specifically, we set the pseudo label of the instance with the largest positive probability in each positive bag to be one. 
The pseudo label $Q$ obtained now will neither fall into degradation, nor violate the rule that there must exist positive instances in each positive bag. 

\subsubsection{Instance-level classifier training and iterative optimization}
After obtaining the pseudo labels of the instances in the positive bags, the instance classifier is trained end-to-end jointly with these instances and the instances in the negative bags. 
As shown in Figure \ref{fig3}, in each iteration, a batch of instances is randomly selected from all instances to train the network, where $x^{n}$ and $y^{n}$ are the instances and their labels from negative bags, and $x^p$ and $q^p$ are the instances from positive bags and their pseudo labels. 
We use the cross entropy loss to train the classifier:
\begin{equation}
    \label{eq9}
    \min _{\theta} C E\left(f\left(x^{n}\right), y^{n}\right)+C E\left(f\left(x^{p}\right), q^{p}\right)
\end{equation}
where $\theta$ is the parameters of $f$. 
After the classifier is trained for a few iterations, the pseudo labels are updated with the new classifier. 
The alternate iterative optimization of network parameters and pseudo labels ensure that the network converges to a local optimal solution. \cite{ref11}. 
The iterative process is intuitively shown in Figure \ref{fig4}. 
At the beginning, the classifier is initialized randomly, so the classification ability of the network is poor, and the pseudo labels generated are mostly wrong. 
However, with the guidance of negative instances from the negative bags and the constraints derived from weak positive bag labels, the pseudo labels will gradually approach the true labels and the network can gradually learn to correctly distinguish the instances, which is demonstrated in experiments.

In addition, in Equation \ref{eq5} the parameter $\mu$ represents the expected positive instance ratio in all the instances in positive bags. 
If $\mu$ is set to a small value, there will only be a small number of positive pseudo labels in each iteration and in the initial iterations the instances with true positive labels will be even fewer because the network’s predictive capability is still very limited at the beginning. 
This will make the network fall into a poor local optimal solution. 
Therefore, we adopt an adaptive $\mu$ parameter as follows:
\begin{equation}
    \label{eq10}
    \mu_{t}=\left\{\begin{array}{c}
    0.5+\frac{\mu-0.5}{T} \cdot t, t<T \\
    \mu, t \geq T
    \end{array}\right.
\end{equation}
where $\mu_{t}$ represents the value of $\mu$ in epoch $t$ and $T$ is the hyperparameter epoch after which $\mu_{t}$ is set to $\mu$. 
In the initial training, $\mu_{t}$ is close to 0.5 to generate an excess of positive pseudo labels, so there will be sufficient true positive labels. 
$\mu_{t}$ gradually approach $\mu$ during iteration.

In inference, all instances are fed into the classifier to obtain instance-level prediction, and the maximum probability of all instances in a bag are contributed to the bag-level prediction:
\begin{equation}
    \label{eq11}
    \hat{Y}_{i}=\max _{j} f\left(x_{i, j}\right)
\end{equation}

\begin{table}[]
  \centering
  \caption{The number of bags in the training and test sets on MNIST-Normal-Bag under different positive instance ratios.}
  \setlength{\tabcolsep}{0.4mm}{
  \begin{tabular}{c|c|c|c|c|c|c}
  \hline
  Positive instance ratio      & 1\%    & 5\%    & 10\%   & 20\%   & 50\%   & 70\%   \\ \hline
  Number of bags (train: test) & 542:90 & 553:91 & 568:94 & 594:98 & 236:40 & 168:28 \\ \hline
  \end{tabular}
  }
  \label{table1}
\end{table}

\begin{figure}[!htbp]
  \begin{center}
  \centerline{\includegraphics[width=1.0\columnwidth]{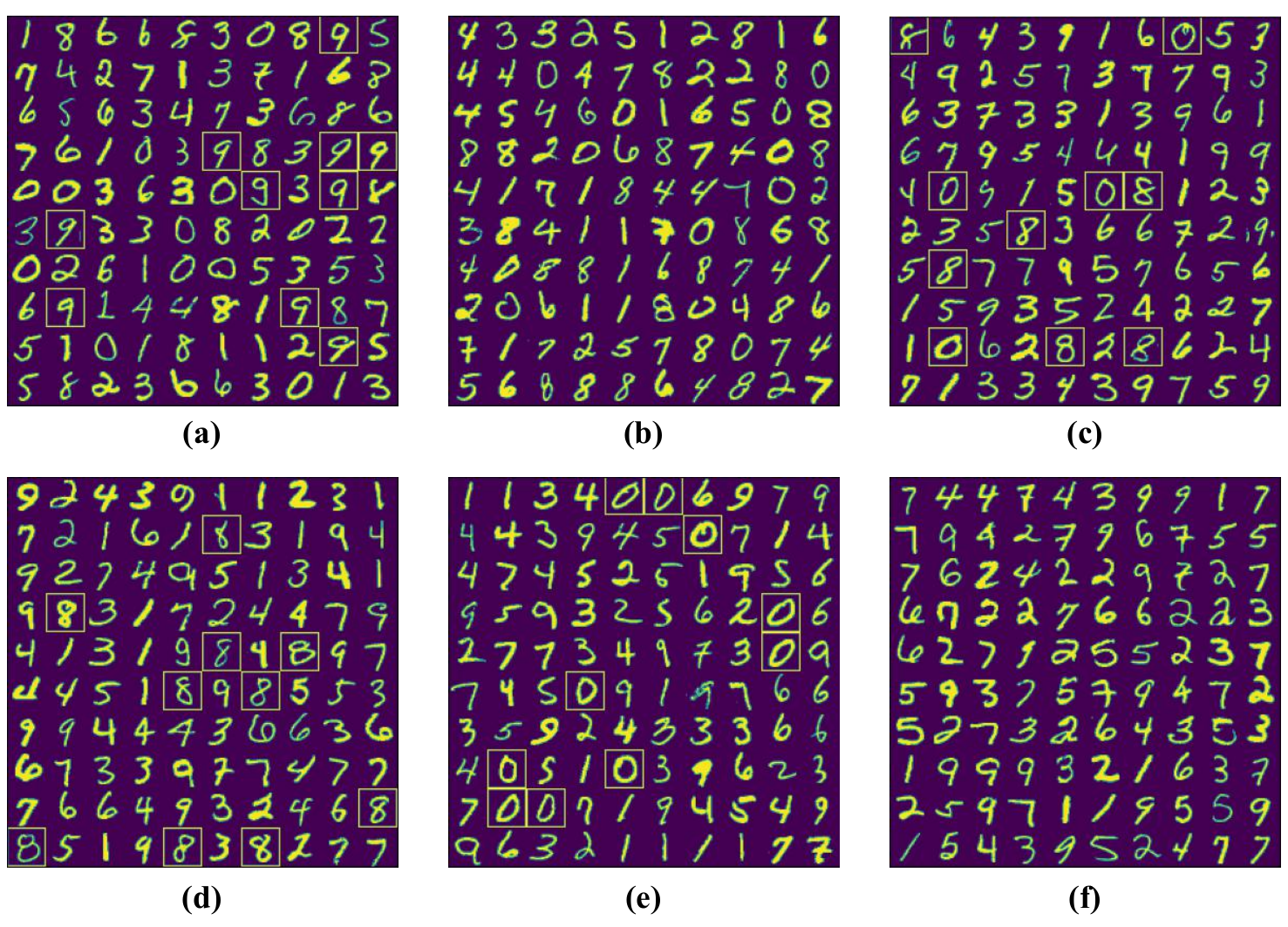}}
  \caption{
    Examples of synthetic bag, where positive instances are mark by yellow boxes: (a) Positive bag in the Synthetic MNIST Normal Bag with a positive instance ratio of 10\%; 
    (b) Negative bag in the Synthetic MNIST Normal Bag; (c) Positive bag in training set of Synthetic MNIST Hard Bag, both number “0” and “8” are positive; 
    (d) Positive bag in test set test-pos8 of Synthetic MNIST Hard Bag, only number “8” appears in the test-pos8; 
    (e) Positive bag in test set test-pos0 of Synthetic MNIST Hard Bag, only number “0” appears in the test-pos0; 
    (f) Negative bag in Synthetic MNIST Hard Bag.
  }
  \label{fig5}
  \end{center}
\end{figure}

\section{Experiments and results}
\subsection{Datasets and evaluation metrics}
In order to demonstrate the problems of existing methods and show the advantages of our proposed method, we first constructed two synthetic MIL tasks using MNIST \cite{ref17}. 
Then we compare our method with a series of SOTA bag-classification and instance-classification methods on five classical MIL datasets and two histopathology image datasets CAMELYON16 and TCGA. 

\noindent
\textbf{Synthetic MNIST Normal Bag.} The MNIST dataset is composed of 70,000 images of handwritten digits with a size of $28 \times 28$. 
To simulate the MIL scenario, we treat each image in MNIST as an instance and label the instances of number "9" as positive and the other numbers as negative. 
We randomly select $a$ positive instances and $100-a$ negative instances from all images without replacement to form a positive bag, whose positive instance ratio is $a/100$. 
Similarly, 100 negative instances are randomly selected without replacement to form a negative bag. 
This process is repeated until all the positive or negative instances in MNIST are used. 
We constructed the MNIST-Normal-Bag dataset with six different positive instance ratios, 1\%, 5\%, 10\%, 20\%, 50\%, and 70\%. 
The number of bags constructed for each positive instance ratio is shown in Table \ref{table1}. 
Visualization of bags are in Figure \ref{fig5}(a) and \ref{fig5}(b).

\noindent
\textbf{Synthetic MNIST Hard Bag.} We construct this dataset to simulate the scenario in which positive instances are divided into easy and hard ones, and this synthetic dataset is denoted as MINST-Hard-Bag. 
In this dataset, we treat both number "0" and number "8" as positive instances, and the other numbers are negative instances. Since in this dataset, positive instances may belong to various concepts, we use this dataset to simulate our methods' strength to deal with various bag-formation.
In positive bags of the training set, these two positive instances appear independently and randomly, and the positive instance ratio is kept at 10\%. 
We constructed three test sets named test-normal, test-pos0 and test-pos8, whose negative bags are similar but the positive bags are different. Concretely, the positive bags of the test-normal contains both number "0" and "8", test-pos0 and test-pos8 consists of only number "0" or number "8", respectively. 
Although which of the two numbers “0” and “8” corresponds to hard instance is unknown and depends on the specific network used to process the images, the latter two test sets can be regarded as special cases of containing only hard positive instance or easy positive instances in positive bags. 
Visualization of bags are in Figure \ref{fig5}(c), \ref{fig5}(d), \ref{fig5}(e), and \ref{fig5}(f).

\noindent
\textbf{Classical MIL Benchmark.} This benchmark consists of five datasets MUSK1, MUSK2, FOX, TIGER and ELEPHANT, and each instance in these datasets is already encoded into a feature vector. 
MUSK1 and MUSK2 are used to predict drug effects, each bag of which contains the molecule with multiple different conformation, but only some of the conformation bind to the target binding site and have drug effects \cite{ref34}. 
The bag that contains at least one effective conformation is labeled positive, otherwise negative. FOX, TIGER and ELEPHANT are image datasets, and the bag that contains at least one animal of interest is labeled positive, otherwise negative \cite{ref45}.

\begin{figure}[!htbp]
  \begin{center}
  \centerline{\includegraphics[width=1.0\columnwidth]{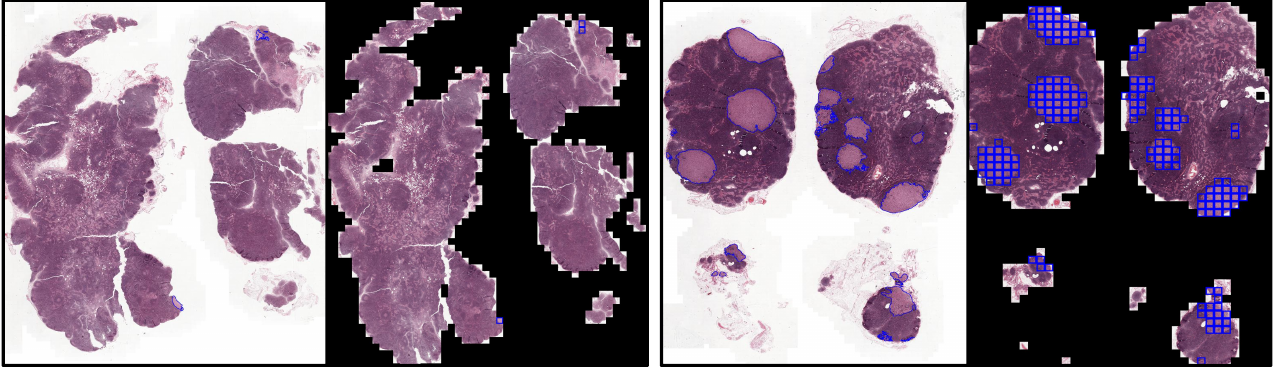}}
  \caption{
  Visualization of two examples in CAMELYON16 dataset. 
  The first and the third images are original positive slides with different sizes of positive areas, which are outlined by blue lines. The second and the forth images are the patches cropped from them, where the background patches have been dropped out and positive patches are marked by blue boxes. 
  }
  \label{fig6}
  \end{center}
  \vspace{-2em}
\end{figure}

\begin{figure}[!htbp]
  \begin{center}
  \centerline{\includegraphics[width=1.0\columnwidth]{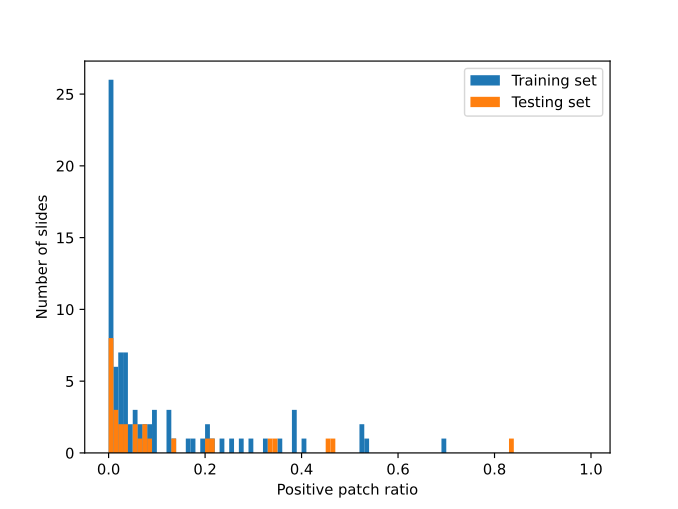}}
  \caption{
    Positive patch ratio distribution of CAMELYON16.
  }
  \label{fig7}
  \end{center}
  \vspace{-1em}
\end{figure}

\noindent
\textbf{CAMELYON16.} CAMELYON16 is a public histopathology image dataset containing 400 whole slide image (WSI) images of lymph nodes (270 for training and 130 for testing) for detecting whether breast cancer metastasis is contained in lymph nodes \cite{ref18}. 
The WSI images that contains metastasis are positive, and the WSI images that do not contain metastasis are negative. The dataset provides both slide label and pixel-level labels of metastasis areas. 
Following similar studies \cite{ref1,ref2,ref6}, to simulate practical histopathology image analysis in weakly supervised scenarios, we only used slide-level labels for training, and the pixel-level labels of cancer areas were only used to test the instance-level classification ability. 
As shown in Figure \ref{fig6}, we divide each WSI into $512 \times 512$ image patches without overlapping under $10\times$ magnification, and patches with an entropy less than 5 are dropped out as background. 
A patch is labeled as positive only if it contains 25\% or more cancer areas; otherwise, it is labeled as negative. 
Eventually, a total of 164,191 patches were obtained, of which there were only 8117 positive patches (4.3\%). 
The number of positive patches in different positive slides varied greatly, and some positive slides contained only several positive patches. To formulate the WSI analysis as a MIL problem, patches are considered as instances and all patches coming from the same slide form a bag. We counted the positive patch ratios of all positive slides, and the distribution is shown in the Figure \ref{fig7}. 
It can be seen that the positive patch ratio is widely distributed, and most of the bags have a positive patch ratio less than 0.01.

\noindent
\textbf{TCGA.} The Cancer Genome Atlas (TCGA) lung cancer dataset is a public histopathology image dataset from National Cancer Institute Data Portal \cite{ref1}. 
The dataset includes 1054 WSIs of two sub-types of lung cancer, lung adenocarcinoma and lung squamous cell carcinoma. 
Following \cite{ref1}, we randomly split the WSIs into 840 training slides and 210 testing slides. After cropping each slide into $224 \times 224$ patches and discarding background patches, the dataset yields 5.2 million patches at $20\times$ magnification. 
Different from CAMELYON 16, the bags in this dataset have a large positive instance ratio (\textgreater 80\% per slide). This dataset only supports bag-level classification task, and we use the pre-extracted instance features provided by \cite{ref1}.

\noindent
\textbf{Evaluation metrics.} For the datasets with instance labels, we report both bag-level classification AUC and instance-level classification AUC. 
For the datasets with only bag labels, we only report bag-level classification AUC or classification accuracy. 
For CAMELYON16 dataset, we also report the FROC score which is defined as the average sensitivity at 6 predefined false positive rates: 1/4, 1/2, 1, 2, 4, and 8 FPs per whole slide image.

\subsection{Implementation details}
For the synthetic datasets from MNIST and the CAMELYON16 dataset, we use AlexNet \cite{ref25} and ResNet18 \cite{ref23}, respectively, for feature extraction, which are followed by a fully-connected layer for classification. 
For the MIL benchmark and TCGA, we only use a fully-connected layer for classification because pre-extracted instance features are provided. 
No network parameters are pretrained. For TCGA, as most patches in positive slides are positive, we replace max-pooling in Equation \ref{eq11} with mean-pooling to have more robust bag predictions. 
The SGD optimizer is used to optimize the network parameters, and the learning rate is fixed at 0.001. 
Because the memory limits of a single NVIDIA 2080Ti GPU, the batch size is 64 for CAMELYON16. For the synthetic dataset, we also use a batch size of 64 for consistency. 
For MIL benchmark and TCGA, as the features have been pre-extracted, we used a batch size of 512. 
For the parameters $\mu$ and $T$ in Equation \ref{eq10}, we directly use the ground truth $\mu$ on the synthetic datasets and use a grid search strategy to find a satisfactoryl $T$ over the grid of $\{50,100,200,400\}$.
On the real datasets MIL benchmark, CAMELYON16 and TCGA, we use a grid search strategy for both $\mu$ and $T$ over the grids $\{0.1,0.15,0.2,0.25\}$ and $\{5,10,20,40\}$, respectively.

\begin{table}[ht]
  \centering
  \caption{Instance-level classification AUC on MNIST-Normal-Bag datasets.}
  \setlength{\tabcolsep}{0.8mm}{
  \begin{tabular}{ccccccc}
  \hline
  Positive instance ratio          & 1\%             & 5\%             & 10\%            & 20\%            & 50\%            & 70\%            \\ \hline
  ABMIL \cite{ref4}          & 0.8483          & 0.9464          & 0.9333          & 0.9490          & 0.9017          & 0.8476          \\
  DSMIL \cite{ref1}         & 0.5150          & 0.9888          & 0.9671          & 0.9385          & 0.9274          & 0.9470          \\
  Loss-Attention \cite{ref3} & 0.7923          & 0.9758          & 0.9727          & 0.9745          & 0.9626          & 0.9416          \\
  Chikontwe \cite{ref2}    & 0.6000          & 0.6677          & 0.9516          & 0.9764          & 0.8884          & 0.8763          \\
  Ours                             & \textbf{0.9745} & \textbf{0.9924} & \textbf{0.9965} & \textbf{0.9962} & \textbf{0.9995} & \textbf{0.9986} \\ \hline
  \end{tabular}
    }
  \label{table2}
\end{table}

\begin{table}[ht]
  \centering
  \caption{Bag-level classification AUC on MNIST-Normal-Bag datasets.}
  \setlength{\tabcolsep}{0.8mm}{
  \begin{tabular}{ccccccc}
  \hline
  Positive instance ratio          & 1\%             & 5\%             & 10\%            & 20\%            & 50\%            & 70\%            \\ \hline
  ABMIL \cite{ref4}         & 0.7232          & \textbf{1.0000} & \textbf{1.0000} & \textbf{1.0000} & \textbf{1.0000} & \textbf{1.0000} \\
  DSMIL \cite{ref1}         & 0.4607          & \textbf{1.0000} & \textbf{1.0000} & \textbf{1.0000} & \textbf{1.0000} & \textbf{1.0000} \\
  Loss-Attention \cite{ref3} & 0.5393          & \textbf{1.0000} & \textbf{1.0000} & \textbf{1.0000} & \textbf{1.0000} & \textbf{1.0000} \\
  Chikontwe \cite{ref2}    & 0.5460          & 0.5590          & \textbf{1.0000} & \textbf{1.0000} & \textbf{1.0000} & \textbf{1.0000} \\
  Ours                             & \textbf{0.9072} & \textbf{1.0000} & \textbf{1.0000} & \textbf{1.0000} & \textbf{1.0000} & \textbf{1.0000} \\ \hline
  \end{tabular}
  }
  \label{table3}
\end{table}

\begin{table}[ht]
  \centering
  \caption{Instance-level and bag-level classification AUCs on synthetic MNIST-Hard-Bag dataset.}
  \setlength{\tabcolsep}{1.5mm}{
  \begin{tabular}{cccc}
  \hline
  Methods                                 & Test set    & Instance-level AUC & Bag-level AUC   \\ \hline
  \multirow{3}{*}{ABMIL \cite{ref4}}          & test-normal & 0.5808             & 0.9983          \\
                                          & test-pos0   & 0.8675             & \textbf{1.0000}         \\
                                          & test-pos8   & 0.5767             & 0.4617          \\ \hline
  \multirow{3}{*}{Loss-Attention \cite{ref3}} & test-normal & 0.6686             & \textbf{1.0000}         \\
                                          & test-pos0   & 0.9685             & \textbf{1.0000}          \\
                                          & test-pos8   & 0.4902             & 0.687           \\ \hline
  \multirow{3}{*}{DSMIL \cite{ref1}}         & test-normal & 0.6928             & \textbf{1.0000}          \\
                                          & test-pos0   & 0.9818             & \textbf{1.0000}         \\
                                          & test-pos8   & 0.4031             & 0.5641          \\ \hline
  \multirow{3}{*}{Chikontwe \cite{ref2}}      & test-normal & 0.7556             & 0.9233          \\
                                          & test-pos0   & 0.6424             & 0.5778          \\
                                          & test-pos8   & 0.8477             & 0.9878          \\ \hline
  \multirow{3}{*}{WENO \cite{ref49}}  & test-normal & 0.7220             & \textbf{1.0000}         \\
                                          & test-pos0   & 0.9970             & \textbf{1.0000}       \\
                                          & test-pos8   & 0.4576             & 0.5499          \\ \hline
  \multirow{3}{*}{Ours}                   & test-normal & \textbf{0.9997}    & \textbf{1.0000} \\
                                          & test-pos0   & \textbf{0.9998}    & \textbf{1.0000} \\
                                          & test-pos8   & \textbf{0.9996}    & \textbf{1.0000} \\ \hline
  \end{tabular}
    }
  \label{table4}
\end{table}

\begin{table*}[ht]
  \caption{Bag-level classification accuracy on MIL benchmark.}
  \centering
  \setlength{\tabcolsep}{4mm}{
  \begin{tabular}{cccccc}
  \hline
  Methods        & MUSK1                  & MUSK2                  & FOX                    & TIGER                  & ELEPHANT                                   \\ \hline
  ABMIL  \cite{ref4}        & 0.892 ± 0.040          & 0.858 ± 0.048          & 0.615 ± 0.043          & 0.839 ± 0.022          & 0.868 ± 0.022                              \\
  ABMIL-Gated \cite{ref4}   & 0.900 ± 0.050          & 0.863 ± 0.042          & 0.603 ± 0.029          & 0.845 ± 0.018          & 0.857 ± 0.027                              \\
  Loss-Attention \cite{ref3} & 0.917 ± 0.066          & 0.911 ± 0.063          & 0.712 ± 0.074          & 0.897 ± 0.065          & 0.900 ± 0.069                              \\
  DSMIL       \cite{ref1}   & 0.932 ± 0.023          & 0.930 ± 0.020          & 0.729 ± 0.018          & 0.869 ± 0.008          & 0.925 ± 0.007                              \\
  WENO       \cite{ref49}   & 0.979 ± 0.034          & 0.937 ± 0.075         & 0.676 ± 0.131          & 0.870 ± 0.083          & 0.927 ± 0.053                              \\
  Ours           & \textbf{0.989 ± 0.033} & \textbf{0.950 ± 0.067} & \textbf{0.810 ± 0.058} & \textbf{0.910 ± 0.049} & \multicolumn{1}{r}{\textbf{0.945 ± 0.027}} \\ \hline
  \end{tabular}
  }
  \label{table5}
\end{table*}

\begin{table*}[ht]
  \centering
  \caption{Instance-level classification and bag-level classification results with CAMELYON16}
  \setlength{\tabcolsep}{5.6mm}{
  \begin{tabular}{ccccc}
  \hline
  Methods                          & End-to-End     & Instance-level AUC & Localization FROC  & Bag-level AUC   \\ \hline
  Fully supervised                 & \textbf{True}  & \textbf{0.9644}   & \textbf{0.5847} & 0.8621          \\  \hline
  ABMIL \cite{ref4}          & \textbf{False}  & 0.7965          & 0.3470   & 0.8171        \\
  DSMIL \cite{ref1}          & \textbf{False}  & 0.8364          & 0.4016   & 0.8343         \\
  Loss-Attention \cite{ref3} & \textbf{False}  & 0.8254          & 0.4044   & 0.8524         \\
  TransMIL \cite{ref6} & \textbf{False}  & -          & -   & 0.8360          \\
  DTFD-MIL  \cite{ref48} & \textbf{False}  & 0.7891           & 0.3607  & 0.8722        \\
  Chikontwe \cite{ref2}    & \textbf{True}  & 0.8581          & 0.3550   & 0.6777         \\ 
  WENO  \cite{ref49}            & \textbf{True}  & 0.9003  & 0.4317 & 0.8352 \\ 
  Ours                             & \textbf{True}  & \textbf{0.9385}  & \textbf{0.4754}  & \textbf{0.9172} \\ \hline
  \end{tabular}
    }
  \label{table6}
\end{table*}

\subsection{Comparison methods}
We denote our proposed method as MIL-SSL, and we compare it to a series of recent methods, including bag-classification methods: ABMIL \cite{ref4}, DSMIL \cite{ref1}, Loss-Attention MIL \cite{ref3}, TransMIL \cite{ref6}, DTFD-MIL \cite{ref48} and instance-classification methods proposed by Chikontwe et al. \cite{ref2} and WENO \cite{ref49}. TransMIL is the first to achieve pathological MIL image classification under a Transformer architecture. DTFD-MIL performs bag-level data augmentation by constructing pseudo-bags. DSMIL proposes a new MIL aggregator that captures the relationships between instances using a dual-stream architecture. WENO utilizes a teacher-student distillation dual-branch structure to achieve classification at both the bag and instance level.
As TransMIL and DTFD-MIL are specially designed for WSI processing which mines the relationship between the instances and solves the limited bag problem respectively, we only compare with them on pathological datasets CAMELYON16 and TCGA. 
For experiments on the datasets we built, we reproduced these methods using their published codes. Since CAMELYON16 is the only real dataset with instance-level labels, we also compared MIL-SSL to the fully supervised instance-classification method. 
For a fair comparison, none of the encoder networks were pretrained. For the experiments on all datasets, we performed a grid search on the key hyperparameters in all the comparison methods. 
The learning rate search grid for all methods were set to $\{0.0001,0.0005,0.001,0.005,0.01\}$. 
For \cite{ref2}, the search grid of hyperparameter $k$ is set to $\{2,4,8,16,32,64\}$. 
For \cite{ref1}, the grid of the hyperparameter $\lambda$ in the loss function is set to $\{0.01,0.1,1,10\}$. 

\subsection{Results on MNIST synthetic datasets}
For the synthetic MNIST-Normal-Bag datasets, the instance-level and bag-level classification AUCs are shown in Tables \ref{table2} and \ref{table3}, respectively. 
The instance-level classification task is harder than the bag-level classification task. 
The proposed MIL-SSL outperforms the other methods under all positive instance ratios in the instance-level classification task, with a large margin in most cases. 
In this hard task, MIL-SSL achieves an AUC larger than 0.9900 in all positive instance ratios equal to or larger than 5\%. 
Even under the scenario of 1\% positive instance ratio, the performance of MIL-SSL is still good. In the bag-level classification task, when the positive instance ratio is greater than or equal to 10\%, all methods can achieve a bag-level classification AUC of 1.0.
However, when the positive instance ratio is only 1\%, MIL-SSL still achieve an AUC of 0.9072, which is higher than the other methods.

There is another phenomenon that is worth noting. As the positive instance ratio increases, the bag classification AUC of all methods reaches 1.0 quickly and maintains at it. 
In contrast, in the instance-level classification task the comparing methods achieve the highest AUC at a positive instance ratio in the middle, and their performance decreases as the positive instance ratio further increases. 
We speculate that the reason is that when there are many positive instances in a positive bag, these methods can realize bag classification by recognizing only a few of the most significant positive instances and have no motivation to identify the other positive instances. 

For the synthetic MNIST-Hard-Bag datasets, in which there are easy and/or hard positive instances in the test sets, the instance-level and bag-level classification AUCs are shown in Table \ref{table4}. 
We first analyze the performance of the bag-classification methods ABMIL, Loss-Attention and DSMIL. When there are both easy positive instances and hard positive instances in the test set (test-normal), their bag-level classification AUCs are very high, but instance-level classification AUCs are very low, which is similar to the results on MNIST-Normal-Bag datasets. 
On test-pos0, they achieve high AUCs on both instance-level and bag-level classification tasks, while on test-pos8, they achieve very low AUCs on both tasks. These results indicate that these methods tend to make decision on the basis of easy positive instances and cannot generalize to the scenarios where there are only hard positive instances. 
We also experiment with the instance-classification method of Chikontwe et al. \cite{ref2} and WENO \cite{ref49}, and the results also show that the network only learns to distinguish one kind of positive instances and completely loses the ability to predict another kind of positive instances. 
On the contrary, the proposed MIL-SSL achieve the best instance-level and bag-level AUCs in all three kinds of test sets, indicating that our classifier can learn to distinguish both easy and hard positive instances, and deal with various positive concepts \cite{Cheplygina_2015}.

\subsection{Results on MIL benchmark}
Since the real datasets in the traditional MIL benchmark are small, we adopt 10-fold cross-validation as in \cite{ref1}, and the results are shown in Table \ref{table5}. 
ABMIL-Gated represents ABMIL with gated attention \cite{ref1}. 
Our method achieves SOTA performance in all five datasets, and especially in FOX, our method outperforms previous methods by a large margin. In addition, we also provide the AUC results for our method: MUSK1 (0.9326 ± 0.1106), MUSK2 (0.9573 ± 0.0654), FOX (0.7980 ± 0.0711), TIGER (0.9264 ± 0.0504), and ELEPHANT (0.9585 ± 0.0237).
Since all the instance features have been pre-extracted in these datasets, we believe that the higher performance of our method stems from the effective information utilization and the hard positive instances learning ability. 

\subsection{Results on CAMELYON16}
\begin{figure}[!htbp]
  \begin{center}
  \centerline{\includegraphics[width=1.0\columnwidth]{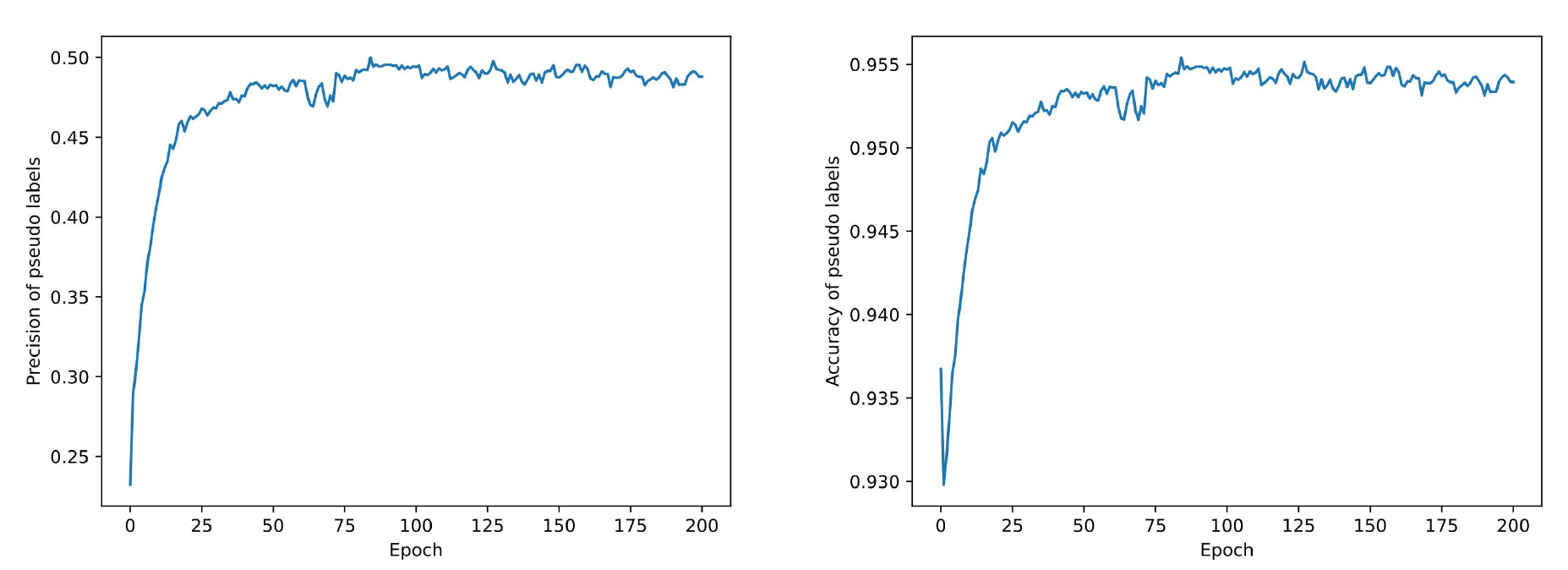}}
  \caption{
    Precision and accuracy of pseudo labels during training.
  }
  \label{fig8}
  \end{center}
\end{figure}

\begin{figure}[!htbp]
  \begin{center}
  \centerline{\includegraphics[width=1.0\columnwidth]{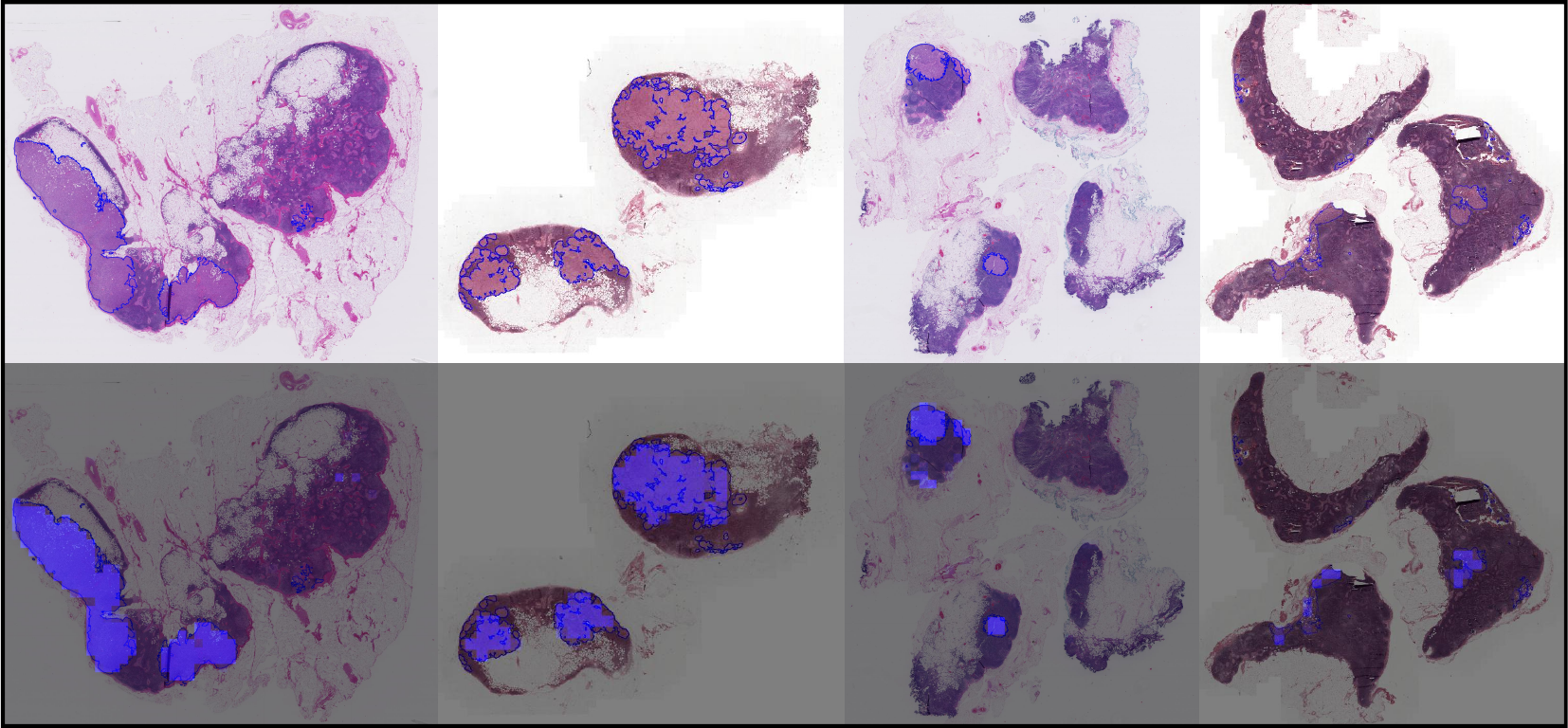}}
  \caption{Visualization of instance-level prediction results on CAMELYON16 test set. 
    In the top row, 4 raw positive WSIs are visualized and their tumor regions are delineated by the blue lines. 
    In the bottom row, instance-level predictions are visualized and deeper blue colors indicate higher probabilities to be tumor.
      }
  \label{fig9}
  \end{center}
  \vspace{-1em}
\end{figure}

\begin{figure}[ht]
\begin{center}
\centerline{\includegraphics[width=1.0\columnwidth]{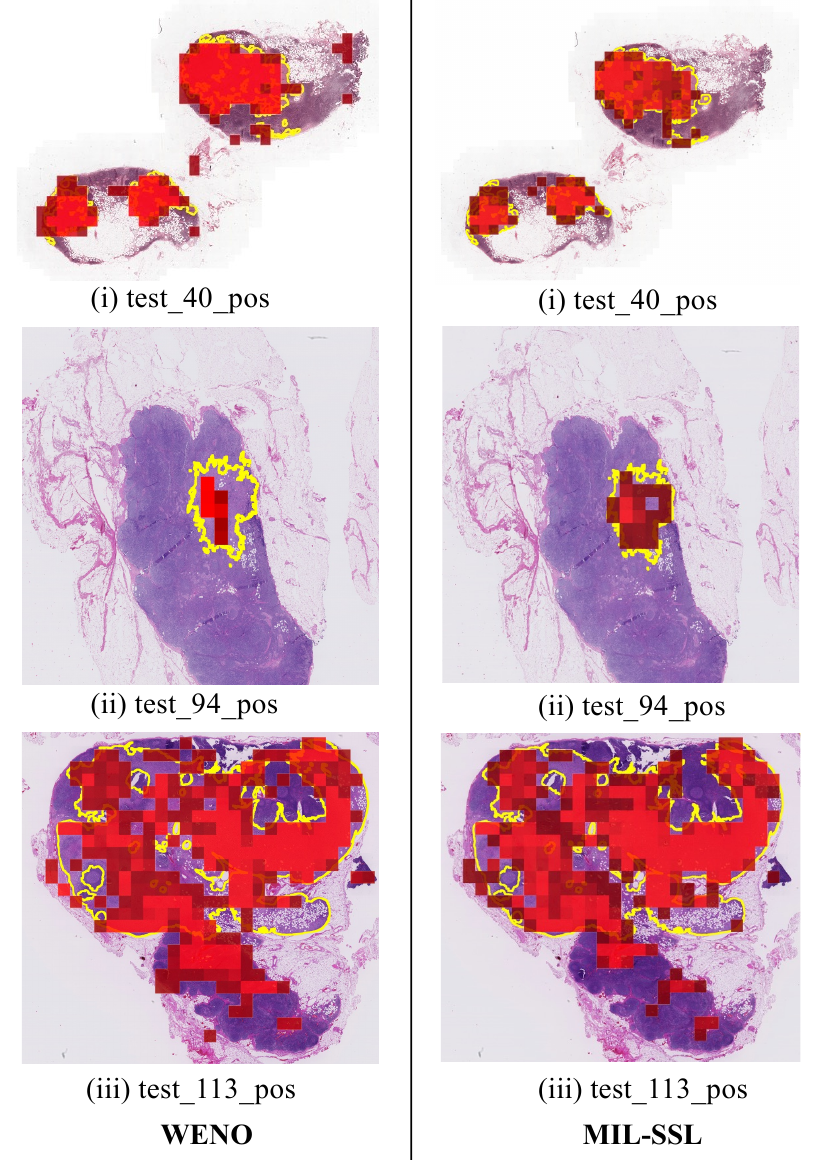}}
\caption{Visualization of instance-level prediction results compared with WENO on CAMELYON16 test set. The model's predictions and the true tumor regions, delineated in yellow. The left column presents results from WENO, while the right column illustrates results from MIL-SSL. MIL-SSL demonstrates superior performance in identifying positive regions and produces fewer false positives.
}
\label{fig10}
\end{center}
\vspace{-1em}
\end{figure}

The instance-level and bag-level classification AUCs are shown in Table \ref{table6} for the real histopathology image dataset CAMELYON16. 
The instance-level classification AUC, Localization FROC and bag-level classification AUC of MIL-SSL are 0.9385, 0.4754 and 0.9172, respectively, which surpass the other methods by a large margin. 
The FROC also suggests our method learn more hard positive instances than others. 
Instance-level prediction results are also visualized in Figure \ref{fig9}.  Furthermore, to demonstrate the weakness of bag-level methods in instance classification, we present the visual fine-grained differences between the WENO method and our method on the CAMELYON16 test set. As shown in Figure \ref{fig10}, our method identifies a greater number of correct instances and has fewer false positives.
In this experiment, ‘Fully supervised’ indicates training an instance classifier directly with the instance labels provided in this dataset, and the bag level prediction is obtained in the same way as MIL-SSL. 
The instance-level AUC of MIL-SSL is only 0.0259 lower than that of ‘Fully supervised’, and the bag-level AUC of MIL-SSL is even significantly higher. 
The precision and accuracy of the generated pseudo labels during training are shown in Figure \ref{fig8}, and it is easy to see that the pseudo labels indeed gradually approach to the true labels.

\begin{table}[ht]
  \caption{Bag-level classification results with TCGA.}
  \centering
  \setlength{\tabcolsep}{8mm}{
  \begin{tabular}{cc}
  \hline
  Methods      & Bag-level AUC   \\ \hline
  Mean-pooling & 0.9369          \\
  Max-pooling  & 0.9014          \\
  MIL-RNN  \cite{ref7}    & 0.9107          \\
  ABMIL  \cite{ref4}      & 0.9488          \\
  DSMIL   \cite{ref1}     & 0.9633          \\
  TransMIL  \cite{ref6}      & 0.9830          \\
  DTFD-MIL  \cite{ref48}      & 0.9808          \\
  WENO    \cite{ref49}    & 0.9727          \\
  Ours         & \textbf{0.9839} \\ \hline
  \end{tabular}
  }
  \label{table7}
\end{table}

\subsection{Results on TCGA}
The bag-level classification results on the TCGA dataset are shown in Table \ref{table7}. 
All comparing methods’ results are from \cite{ref1}. 
The Mean-pooling and Max-pooling are traditional bag-classification approaches, in which bag features are obtained simply by mean-pooling and max-pooling. 
MIL-RNN \cite{ref7} is a bag-classification approach with an RNN aggregator. 
This bag classification task is relatively easy because the positive instance ratio within the positive bags is high in this dataset, and our method still achieves the highest AUC of 0.9839.

\subsection{Ablation study}
We further conduct ablation study on the key components of MIL-SSL using the CAMELYON16 dataset. 
The results are shown in Table \ref{table8}. 
‘Soft pseudo label’ means that the generated pseudo labels are real numbers in (0, 1) and without soft pseudo label, the generated pseudo labels are binarized into 0 or 1. 
‘Constrain’ represents whether to constrain the positive instance ratio on the generated pseudo labels and we directly use Formula \ref{eq2} to generate the pseudo labels if the constraining is not used. 
When adaptive $\mu$ is not used, we directly fix the parameter $\mu$ to 0.15. 
The ablation study results show that the soft pseudo label and the constraint on the positive ratio of the pseudo labels play key roles in the effectiveness of MIL-SSL. 
Without these two techniques, the network can hardly distinguish whether an instance is positive or not, and we also observe that all the pseudo labels are assigned negative in this case. 
After using both soft pseudo labels and the constraints, the instance-level and bag-level classification AUCs reach 0.9073 and 0.7995, respectively. 
Though the instance-level AUC already surpass existing methods listed in Table \ref{table6}, the bag-level AUC is still poor because the positive instance ratio is very low in CAMELYON16 and MIL-SSL falls into bad local optimal solution. 
Finally, with the adaptive $\mu$ strategy used, MIL-SSL is able to find better solution and achieve the state-of-the-art results. 

\begin{table}[ht]
  \caption{Ablation study results on CAMELYON16 dataset.}
  \centering
  \setlength{\tabcolsep}{0.4mm}{
  \begin{tabular}{ccccc}
    \hline
  Soft pseudo label & Constrain & Adaptive $\mu$ & Instance-level AUC & Bag-level AUC \\ \hline
    &   &   & 0.5292          & 0.5513          \\
  $\checkmark$ &   &   & 0.8339          & 0.6807          \\
  $\checkmark$ & $\checkmark$ &   & 0.9073          & 0.7995          \\
  $\checkmark$ & $\checkmark$ & $\checkmark$ & \textbf{0.9385} & \textbf{0.9172} \\ \hline
  \end{tabular}
  }
  \label{table8}
\end{table}

\subsection{Robustness to hyperparameter $\mu$}
Finally, we study the robustness of our algorithm to the main hyperparameter $\mu$ on CAMELYON16. 
The dynamic $\mu$ parameter is still used in this experiment. 
Concretely, at the beginning of training, $\mu_t$ is 0.5, and then gradually decreases to a stable value $\mu$. 
We experiment on four $\mu$ values, 0.25, 0.20, 0.15 and 0.10, and the results are shown in Table \ref{table9}. 
We can see that the performance is fairly robust to different choices of $\mu$ values. 
An optimal $\mu$ value can be easily found by grid search.

\begin{table}[ht]
  \centering
  \caption{Instance-level classification and bag-level classification results of MIL-SSL with different $\mu$ in CAMELYON16.}
  \setlength{\tabcolsep}{4mm}{  
  \begin{tabular}{cccc}
  \hline
          & $\mu$   & Instance-level AUC & Bag-level AUC   \\ \hline
  MIL-SSL & 0.25 & 0.9057             & 0.8851          \\
  MIL-SSL & 0.20 & 0.9200             & 0.8906          \\
  MIL-SSL & 0.15 & \textbf{0.9385}    & \textbf{0.9172} \\
  MIL-SSL & 0.10 & 0.8806             & 0.8947          \\ \hline
  \end{tabular}
  }
  \label{table9}
\end{table}

\section{Conclusion}
In this paper, we formulate multiple instance learning as a semi-supervised instance classification problem and propose a weakly-supervised self-training method to learn from both labeled and unlabeled instances. 
In our method, both the labeled and the unlabeled instances can be effectively utilized to train a better instance-level classifier, on the base of which a better bag-level classifier is also obtained. 
Extensive experiments on both synthetic and real datasets demonstrate the superiority of the proposed method. 
Furthermore, since our method is based on semi-supervised learning, unlabeled bag data can also be directly used to further improve the model performance, which could be a future research direction.

\bibliographystyle{elsarticle-num}

\bibliography{egbib}
\begin{IEEEbiography}[{\includegraphics[width=1in,height=1.25in,clip,keepaspectratio]{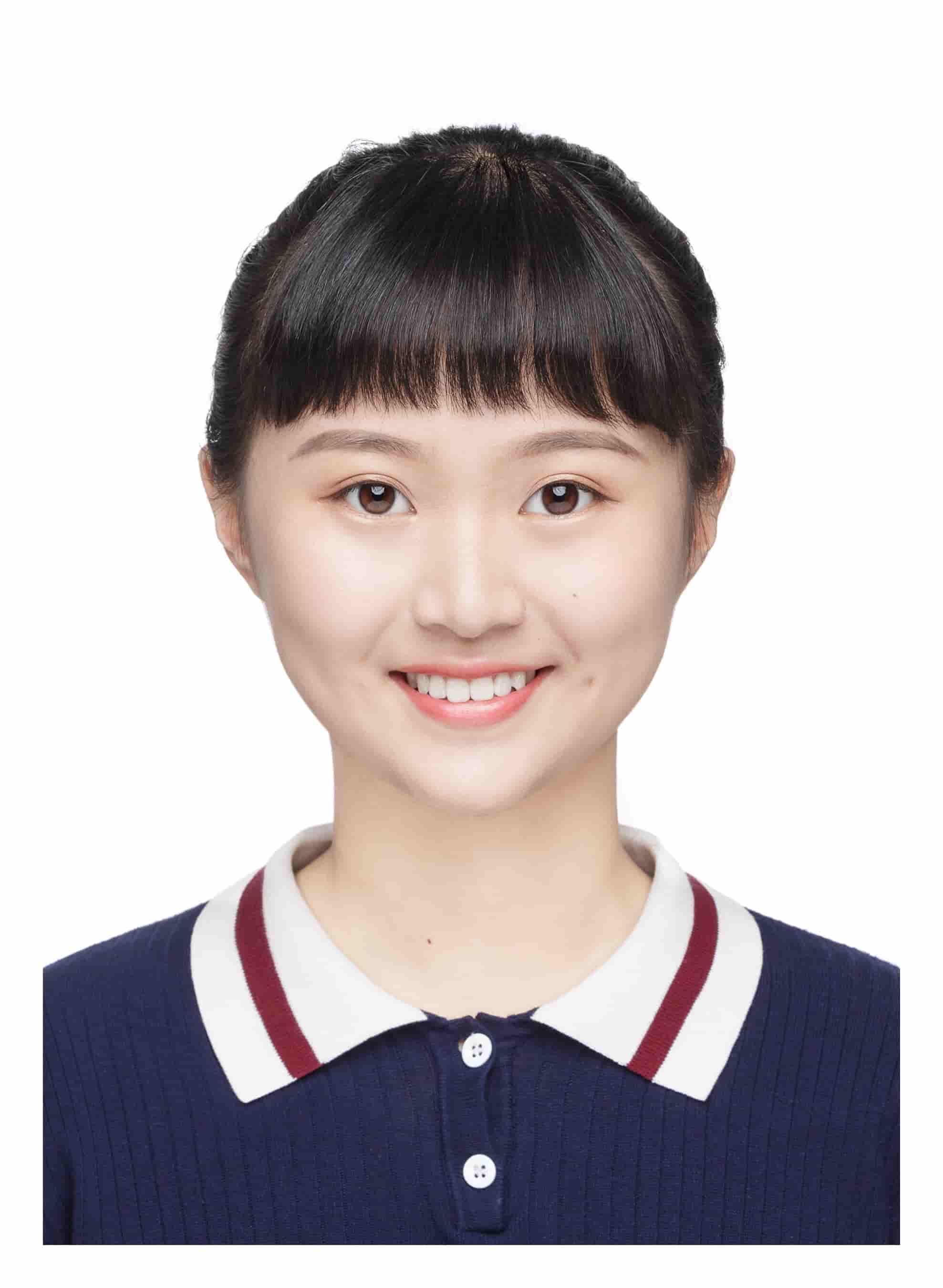}}]{Yingfan Ma}
received the B.S. degree in computer science from the Harbin Institute of Technology(HIT) in 2022. She is currently pursuing the Master degree in biomedical engineering from the Digital Medical Research Center, School of Basic Medical Sciences, Fudan University, Shanghai, China. Her research interests include computer vision and data mining.

  \end{IEEEbiography}
\begin{IEEEbiography}[{\includegraphics[width=1in,height=1.25in,clip,keepaspectratio]{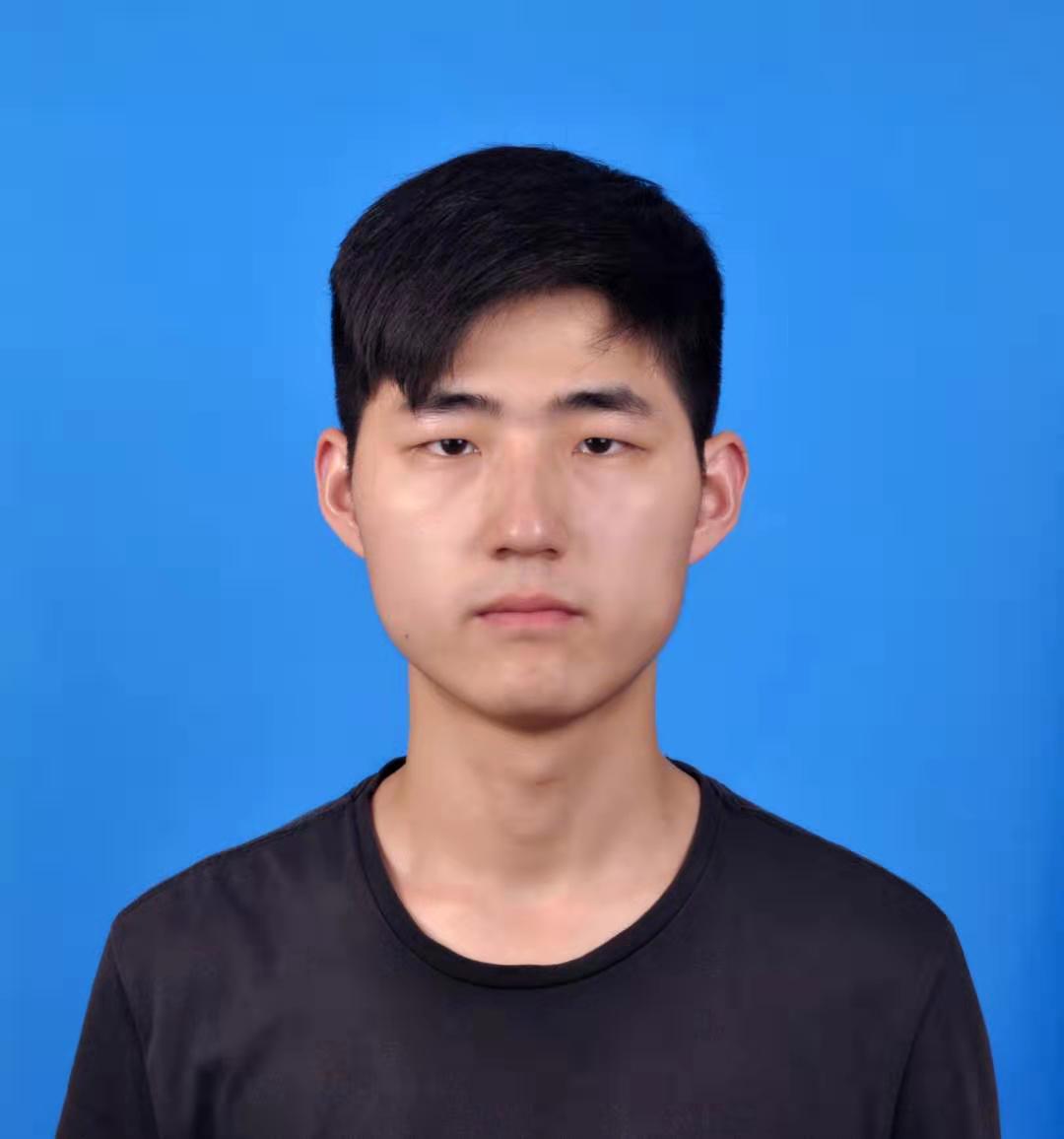}}]{Xiaoyuan Luo}
    received the B.S. degree in automation from Wuhan University of Technology, Wuhan, China, in 2019. 
    He is currently a Ph.D. student in School of Basic Medical Science of Fudan University. 
    His research interests include medical image processing and computer vision.
  \end{IEEEbiography}
  
  \begin{IEEEbiography}[{\includegraphics[width=1in,height=1.25in,clip,keepaspectratio]{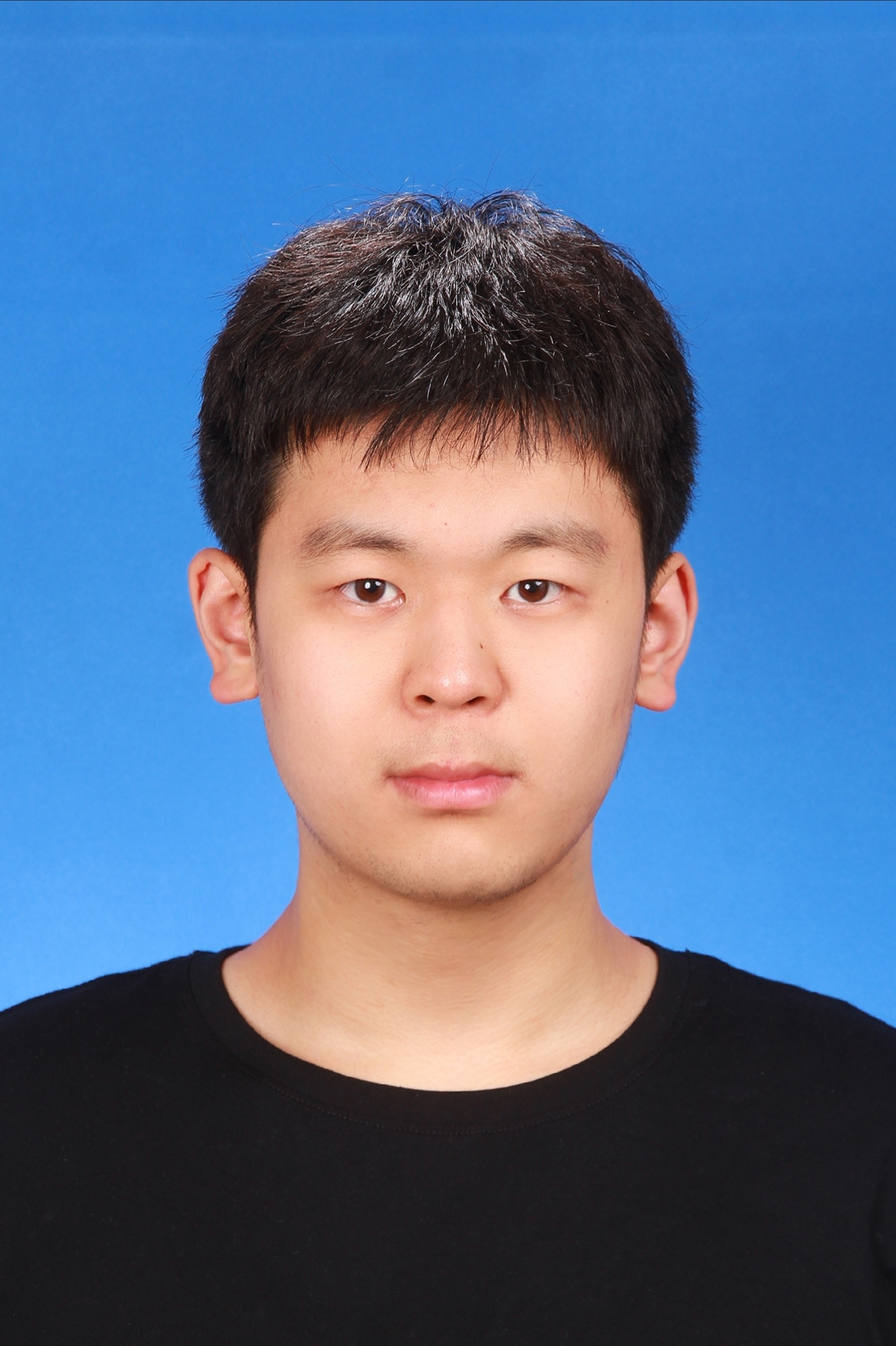}}]{Mingzhi Yuan}
    received the B.S. degree in communication engineering from the Harbin Institute of Technology (HIT) in 2020. 
    He is currently a Ph.D. student in School of Basic Medical Science of Fudan University. 
    His research interests include 3D vision and medical image processing. 
  \end{IEEEbiography}
  
  \begin{IEEEbiography}[{\includegraphics[width=1in,height=1.25in,clip,keepaspectratio]{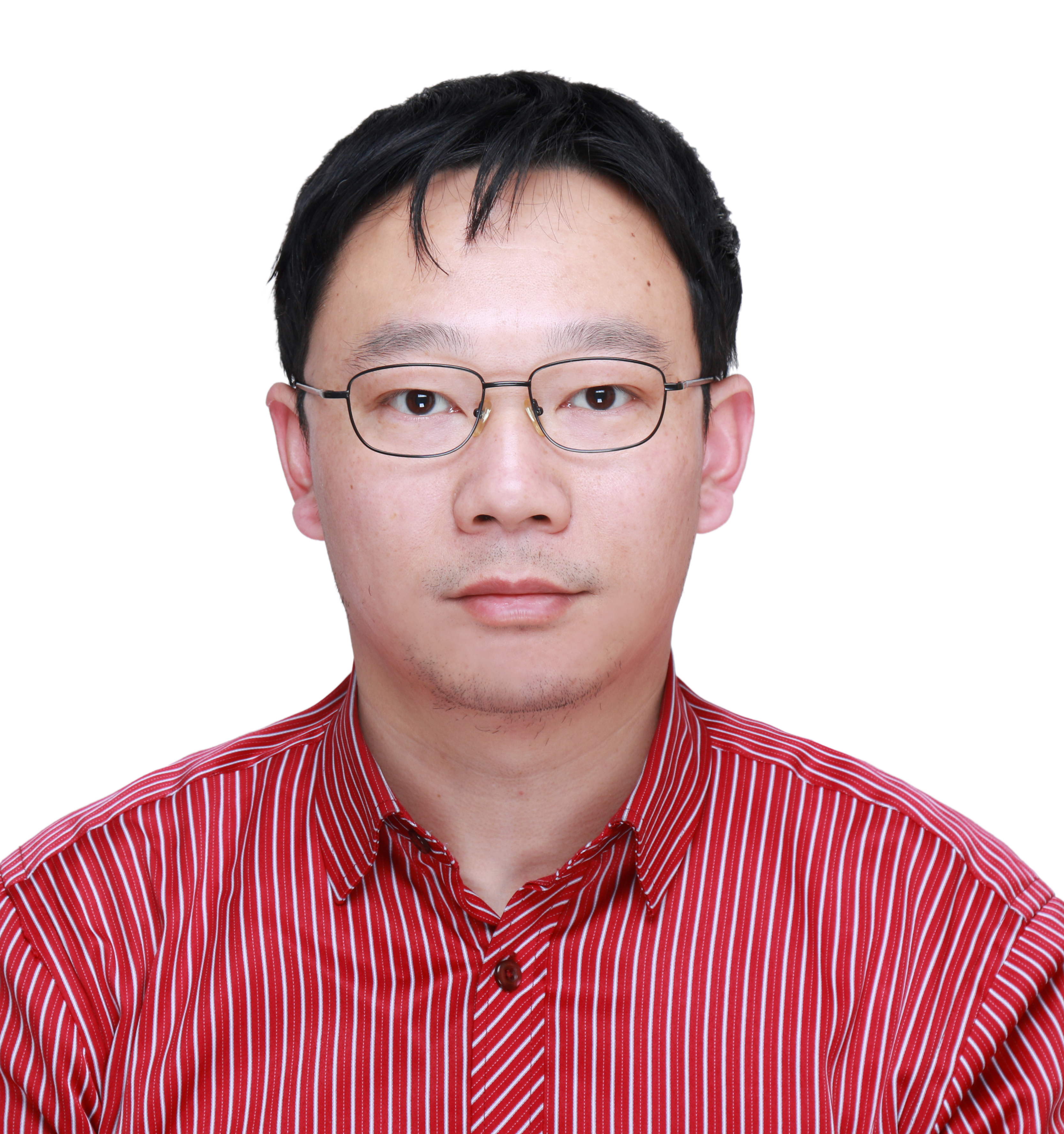}}]{Xinrong Chen}
    received the B.S. degree in electronic and information engineering from the Nanjing University of Science and Technology, in 2004, and the M.S. degree in signal and information processing from Southeast University, Nanjing, China, in 2007, and the Ph.D. degree in biomedical engineering from Fudan University, China, in 2014. 
    He was with the Institute of Neuroscience and Medicine -4, Forschungszentrum Juelich, Germany, as a Postdoctoral Researcher, from 2016 to 2018. 
    He is currently a Professor with the Academy for Engineering and Technology, Fudan University, China. 
    His current research interests include medical image analysis, computer vision, and image-guided surgery.
  \end{IEEEbiography}
  
  \begin{IEEEbiography}[{\includegraphics[width=1in,height=1.25in,clip,keepaspectratio]{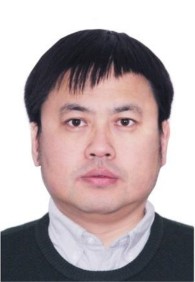}}]{Manning Wang}
    received the B.S. and M.S. degrees in power electronics and power transmission from Shanghai Jiaotong University, Shanghai, China, in 1999 and 2002, respectively. 
    He received Ph.D. in biomedical engineering from Fudan University in 2011. 
    He is currently a professor of biomedical engineering in School of Basic Medical Science of Fudan University. 
    His research interests include medical image processing, image-guided intervention and computer vision. 
  \end{IEEEbiography}

\clearpage

\appendix
\section{Proof of the Proposition 1}\label{proof}

Proposition 1 and the proof are given bellow.

\textbf{Proposition 1:} Given $K>1$ independent and identically distributed instances, the probability of each instance $x_i$ being positive is $P \in(0,1)$, and the information entropy of these instances is $H_I$. 
After the instance aggregation operation is performed on the $K$ instances, and the information entropy of the bag is $H_B$, then $H_{B}<H_{I}$.

The information entropy of instance labels is:
\begin{equation}
    \label{proeq1}
    H_{I}\left(X_{i}\right)=-K(P \log P+(1-P) \log (1-P))
\end{equation}

When only bag label available, the information entropy of bag label is:
\begin{equation}
    \label{proeq2}
    H_{B}\left(X_{i}\right)=-\left(P^{K} \log P^{K}+\left(1-P^{K}\right) \log \left(1-P^{K}\right)\right)
\end{equation}

Then we prove $H_{I}\left(X_{i}\right) \geq H_{B}\left(X_{i}\right)$ and the two sides are equal if and only if $K=1$ or $P=0$ or $P=1$. 

\begin{equation}
    \begin{aligned}
      \text{Set} \ L^{K}(P) &=H_{I}\left(X_{i}\right)-H_{B}\left(X_{i}\right) \\
      & =-K(P \log P+(1-P) \log (1-P))+ \\ 
      & \quad \left(P^{K} \log P^{K}+\left(1-P^{K}\right) \log \left(1-P^{K}\right)\right)
    \end{aligned}
\end{equation}

For $K=1$, $L^{K}(P)=L^{1}(P)=0$.

For $K>1$, 
\begin{equation}
    \begin{aligned}
    &L^{K+1}(P)-L^{K}(P) \\
    &=-(P \log P+(1-P) \log (1-P))+ \\
    &\quad \left(1-P^{K+1}\right) \log \left(1-P^{K+1}\right)+P^{K+1} \log P^{K+1}- \\ 
    &\quad \left(1-P^{K}\right) \log \left(1-P^{K}\right)-P^{K} \log P^{K} \\
    &= \Delta L_{1} + \Delta L_{2}
    \end{aligned}
\end{equation}

\begin{equation}
  \Delta L_{1} = \left(-P+(K+1) P^{K+1}-K P^{K}\right) \log P \\
\end{equation}

\begin{equation}
  \begin{aligned}
  \Delta L_{2} & = -(1-P) \log (1-P) -(1-P^{K}) \log (1-P^{K}) + \\
  &\quad (1-P^{K+1}) \log (1-P^{K+1})
  \end{aligned}
\end{equation}

For $\Delta L_{1}$, the boundary is derived as:
\begin{equation}
  \begin{aligned}
  &-P+(K+1) P^{K+1}-K P^{K} \leq-P+(K+1) P^{K}-K P^{K} \\
  &-P+(K+1) P^{K}-K P^{K} =-P+P^{K} \leq 0 
  \end{aligned}
\end{equation}
and $\log P \leq 0$.

So $\Delta L_{1} \geq 0$ is satisfied, and the two sides are equal if and only if $P=0$ or $P=1$. 

For $\Delta L_{2}$, the boundary is derived as:
\begin{equation}
    \begin{aligned}
    &-(1-P) \log (1-P)+\left(1-P^{K+1}\right) \log \left(1-P^{K+1}\right)-\\
    &\left(1-P^{K}\right) \log \left(1-P^{K}\right) \geq \\
    &-(1-P) \log (1-P)+\left(1-P^{K+1}\right) \log \left(1-P^{K}\right)- \\
    &\left(1-P^{K}\right) \log \left(1-P^{K}\right)= \\
    &-(1-P) \log (1-P)+\left(P^{K}-P^{K+1}\right) \log \left(1-P^{K}\right)= \\
    &-(1-P) \log (1-P)+P^{K}(1-P) \log \left(1-P^{K}\right)= \\
    &(1-P)\left(-\log (1-P)+P^{K} \log \left(1-P^{K}\right)\right) \geq \\
    &(1-P)\left(P^{K} \log (1-P)-\log (1-P)\right)= \\
    &(1-P)\left(P^{K}-1\right) \log (1-P) \geq 0
    \end{aligned}
\end{equation}

So $\Delta L_{2} \geq 0$ is satisfied and the two sides are equal if and only if $P=0$ or $P=1$. 

Now $L^{K+1}(P)-L^{K}(P) \geq 0$ has been proved, and we have $L^{1}(P) = 0$, thus $H_{I}\left(X_{i}\right) \geq H_{B}\left(X_{i}\right)$ is satisfied and the two sides are equal if and only if $K=1$ or $P=0$ or $P=1$. 

Thus for $K>1$ and $P \in(0,1)$, we have the conclusion: $H_{B}<H_{I}$. 



\section{Information entropy difference between instance labels and bag label}\label{figure}
Figures for Information entropy difference between instance labels and bag label is given blow.

\begin{figure}[ht]
  \begin{center}
  \centerline{\includegraphics[width=0.86\columnwidth]{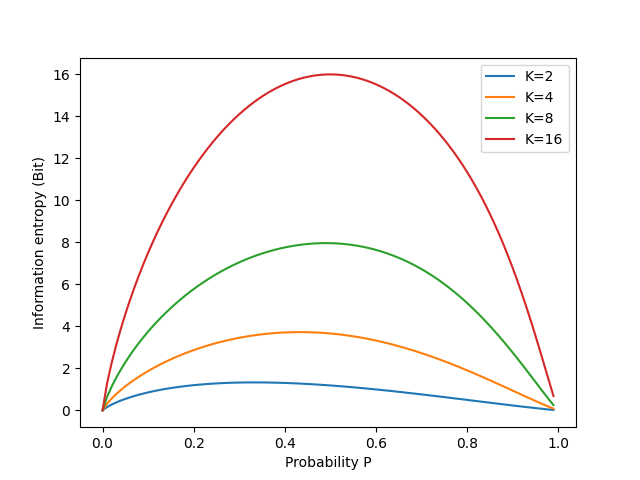}}
  \caption{
    Information entropy difference between instance labels and bag label under different $K$.
  }
  \label{fig2}
  \end{center}
  \vspace{-0.8em}
\end{figure}

\end{document}